\documentclass[letterpaper]{article}
\usepackage{aaai}
\usepackage{times}
\usepackage{helvet}
\usepackage{courier}
\usepackage{pgfgantt}
\usepackage{graphicx}
\usepackage{float}
\usepackage[T1]{fontenc}

\title{Situated Structure Learning of a Bayesian Logic Network for Commonsense Reasoning}
\author{Haley Garrison \and Sonia Chernova\\
Georgia Institute of Technology\\
Atlanta, GA\\
hgarrison3@gatech.edu, chernova@cc.gatech.edu\\}
\copyrightyear{2016}
\begin{document}
\maketitle
\begin{abstract}
\begin{quote}
This paper details the implementation of an algorithm for automatically generating a high-level knowledge network to perform commonsense reasoning, specifically with the application of robotic task repair.  The network is represented using a Bayesian Logic Network (BLN) \cite{Jain2009BayesianNetworks}, which combines a set of directed relations between abstract concepts, including \emph{IsA}, \emph{AtLocation}, \emph{HasProperty}, and \emph{UsedFor}, with a corresponding probability distribution that models the uncertainty inherent in these relations.  Inference over this network enables reasoning over the abstract concepts in order to perform appropriate object substitution or to locate missing objects in the robot's environment.  The structure of the network is generated by combining information from two existing knowledge sources: ConceptNet \cite{Speer2012Representing5}, and WordNet \cite{Miller1995WordNet:English}.  This is done in a "situated" manner by only including information relevant a given context.  Results show that the generated network is able to accurately predict object categories, locations, properties, and affordances in three different household scenarios.
\end{quote}
\end{abstract}
\section{Introduction}
Imagine a world in which autonomous robots are available for everyday people: you could go to the store, pick up a robot, and place it in your home. You could ask the robot to make dinner, do your laundry, or clean the house.  However, in order for a robot to execute such high-level tasks in new or uncertain environments, it must be able to adapt the learned tasks to its local environment and repair any missing information from the tasks. For example, say a robot is cooking a known recipe in a new kitchen. The cookware and other objects it originally used no longer exist. It must instead reason about high level concepts (ex. pots and pans), determine which ones are suitable for the task at hand (ex. an object that can be used as a container), and find those objects in the new kitchen based on knowledge of their likely locations (ex. pans can be found in cabinets).

For this type of abstract reasoning to be possible, the robot must be able to consult a commonsense knowledge network and make inferences over the concepts in this network.  To achieve good performance on such inference tasks, the network must have the following properties:

\begin{enumerate}
\item The network is situated - it contains only information relevant to the current context
\item The size of the network is small enough for fast, online inference
\end{enumerate}

In order to prevent excessive noise in the network and reduce its size, contextually irrelevant concepts and relations must be excluded.  However, even if irrelevant information is excluded, the size of the network may still be too large to facilitate fast inference, so it will also be necessary to exclude concepts and relations that are redundant or carry little valuable information.  Since no existing knowledge network holds both of these properties, an automated structure learning algorithm was implemented to combine information from existing sources in a situated manner based on the current context of the robot.

\begin{figure*}
\centering
\includegraphics[width=\textwidth]{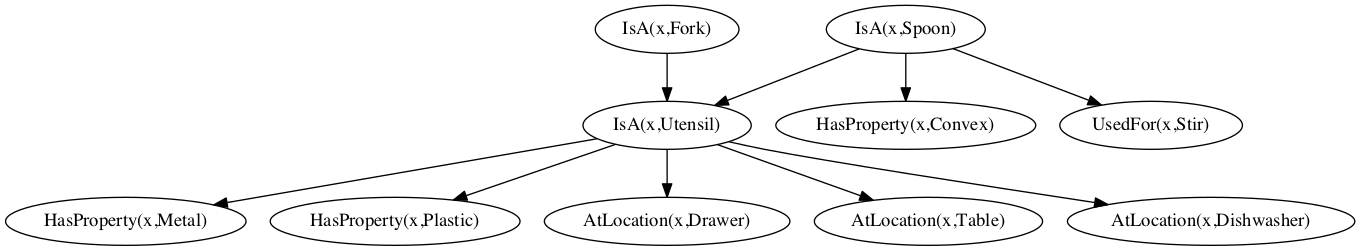}
\caption{An example of high-level knowledge representation using a BLN.}
\label{Sample}
\end{figure*}

\section{Related Work}
Two existing commonsense knowledge networks that are widely used for a variety of applications are WordNet \cite{Miller1995WordNet:English} and ConceptNet \cite{Speer2012Representing5}.  WordNet consists of a collection of synsets, which connect concepts hierarchically through the \emph{IsA} relation.  WordNet also distinguishes between different senses of the same word and provides glosses, or definitions, for each sense.  While WordNet is clean and hand-coded, it also lacks diversity in the types of relations it contains.  ConceptNet, on the other hand, contains a wide variety of different relations, but it does not distinguish between word senses and it is not hand coded, leading to a large amount of noise.

The closest known work to that proposed in this paper is the KnowRob project \cite{Tenorth2013KnowRobRobots}.  In this work, the authors created a knowledge network from a variety of encyclopedic sources and represented the network using Prolog rules and the Web Ontology Language (OWL).  This network was used to perform plan repair by filling in missing low-level details from high-level task descriptions.  However, this representation resulted in a large network without contextual refinement.  It also consisted of many separate components and lacked a unified model.  Furthermore, the concepts used in the network were manually selected according to the perceived relevance to robotic applications rather than automatically generating the network.

Other related works have had similar shortcomings.  Zhu, et al. \cite{Zhu2014ReasoningRepresentation} performed affordance prediction on a set of images by using a Markov Logic Network (MLN) \cite{Richardson2006MarkovNetworks} to represent affordance knowledge.  Like KnowRob, this work did not deal with context and used hand-selected objects and affordances in the network.  In \cite{Chen2011CombiningDisambiguation}, contextual noise was addressed by disambiguating the concepts in ConceptNet to enrich the WordNet senses with more diverse knowledge for improved performance on word sense disambiguation tasks.  While disambiguating ConceptNet helped provide context for each of its concepts, the resulting knowledge base was not further limited in size based on the context of any particular domain.  In contrast to this approach, \cite{Stoica2004Nearly-AutomatedCreation} did construct a situated knowledge hierarchy in a (nearly) automated way.  However, it only included the \emph{IsA} relation and did not enrich this information with other relations from sources like ConceptNet.

\section{Bayesian Logic Networks}
The knowledge network generated by this work is represented using a Bayesian Logic Network (BLN) \cite{Jain2009BayesianNetworks}.  BLN's are a type of directed statistical relational model that serves as a template for a Bayesian Network by representing each node as a function/predicate with arguments rather than a single random variable.  Additionally, BLN's allow logical constraints, represented as first-order logic rules, to be imposed on the network.  A BLN is formally defined as a tuple, $\mathcal{B} = (\mathcal{D}, \mathcal{F}, \mathcal{L})$, such that:
\begin{itemize}
\item $\mathcal{D} = (\mathcal{T}, S, E, t)$ is the declaration, where $\mathcal{T}$ consists of the declared types, $S$ is a set of function signatures, and $E$ is a set of abstract entities where $t: E \rightarrow 2^T \, \backslash \, \{ \emptyset \}$ is a mapping from each entity its respective types.
\item $\mathcal{F}$ defines a set of "fragments" of a conditional probability distribution.  Each fragment represents a directed conditional dependence between two abstract random variables (a parent and a child).  These random variables consist of a function $f(p_1, ..., p_n), f \in S$, where each of the parameters of $f$ can either be a "meta-variable" or an entity $e \in E$.  The fragments are represented by a conditional probability function (CPF) that specifies a distribution over the child variable for each configuration of the parent variables.
\item $\mathcal{L}$ is a set of deterministic constraints described as first-order logic formulas over the abstract random variables.
\end{itemize}

Before inference can be performed on the network, a mixed ground instantiation $M = ((X, D, G, P), (X, D, C))$ of the BLN, $\mathcal{B}$, is generated.  In this case, $X$ is the set of grounded random variables, $f(e_1, ..., e_n), e_i \in E$, $D$ is the domain of the random variables produced by each grounding, $G$ specifies the connectivity of the graph given by the fragments in $\mathcal{F}$, and $P$ is the conditional probability function for each random variable, as determined by $\mathcal{F}$.  The first-order logic formulas in $\mathcal{L}$ are grounded by substituting the abstract random variables with their groundings and applying constraints $C$ that specify the configurations of the random variables required to satisfy the logic formula.

The inference process proceeds on the grounded network by conditioning the query variables, $Q$, on the set of evidence variables, $V$, and marginalizing over the non-query variables, $N$:

\[
P(Q \mid V = v) = \int_{N} P(Q, N \mid V = v) \, \mathrm{d}N 
\]

Since this marginalization grows exponentially with the number of non-query variables, approximate inference algorithms that have been applied to traditional Bayesian Networks, such as Likelihood Weighting \cite{Fung2013WeighingNetworks} or Gibbs Sampling \cite{Geman1984StochasticImages}, can be used as an alternative.

To handle the logical constraints, boolean auxiliary variables are added to $X$ for each constraint in $C$ with parent nodes in $G$ for each of the random variables involved in the constraint.  This allows the inference process described above to remain unchanged with the addition of logical constraints.

Although some similar works such as \cite{Zhu2014ReasoningRepresentation} use undirected Markov Logic Networks (MLN) \cite{Richardson2006MarkovNetworks}, a directed network was chosen for this work because it more explicitly models the directed nature of the relations between concepts.  In preliminary tests, the BLN representation was able to perform complex reasoning in both the causal and diagnostic directions, while the MLN suffered poor performance when trying to reason in both directions.  Furthermore, MLNs require a gradient descent on the pseudo-log-likelihood to learn the network weights.  As a result, the learning process for MLNs is much slower than BLNs which use a simple maximum likelihood frequency count.
% Insert sentence here on why I am not using MLNs
% - Directed network more explicitly models directed nature of relations
% - MLNs are not great for reasoning in both the causal and diagnostic directions
% - MLNs take a long time to learn weights because have to do gradient descent on
%	pseudo-log-likelihood, BLNs just use MLE (frequency count)

\begin{figure*}
\centering
\includegraphics[width=\textwidth]{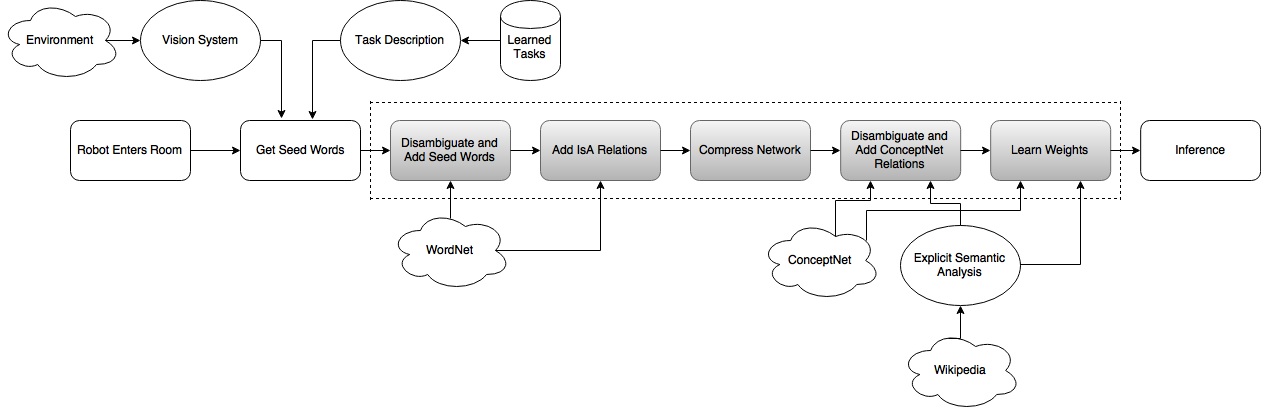}
\caption{A flowchart showing the overall approach taken to generate the network.  The dashed box shows the components of the algorithm, while white boxes are from external sources.}
\label{Netgen}
\end{figure*}

\section{Network Representation}
For the proposed knowledge network, the predicates, $f$, were chosen to be boolean with the following signatures and associated parameter types in $\mathcal{T}$:

\begin{itemize}
\item IsA(object, concept)
\item HasProperty(object, property)
\item AtLocation(object, location)
\item UsedFor(object, affordance)
\end{itemize}

The \emph{IsA}, \emph{HasProperty}, and \emph{UsedFor} relations were chosen because they can be used to perform object substitution for plan repair by finding objects that are similar to the original object, or objects that can perform the same function as the missing object.  The \emph{AtLocation} relation will allow the robot to reason about possible locations of objects to enable it to find missing objects.  For each predicate, the "object" parameter will be a meta-variable that will represent a grounded instance of an object over which the robot can reason, and the "concept," "property," "location," and "affordance" parameters will represent abstract entities.

The general structure of the network fragments in $\mathcal{F}$ will consist of connections such as those shown in Figure \ref{Sample}.  Each of these fragments will be associated with a discrete CPF that represents the likelihood of their occurrence.  For example, utensils may have some likelihood that they are metal versus plastic, and they have some likelihood that they will be found in a drawer compared to a table or a dishwasher.  For the purposes of this paper, no logical constraints were imposed on the network, as experiments so far have shown that high probability relations can effectively be modeled by assigning a probability of one to the corresponding fragment.

\section{Network Generation}
An overview of the approach taken for the network generation can be found in Figure \ref{Netgen}.  The dashed line indicates the components that were implemented as part of the network generation algorithm.

\subsection{Getting Seed Words}
Before the network can be generated, a set of seed words must first be obtained.  These seed words should be related to the domain in which the robot is operating and could come from the robot's vision system (objects it sees in its environment), or from the task description.  For testing purposes, a set of objects from three different household tasks was extracted and used as input to the network generation algorithm.

\begin{figure*}
\centering
\includegraphics[width=0.6\textwidth]{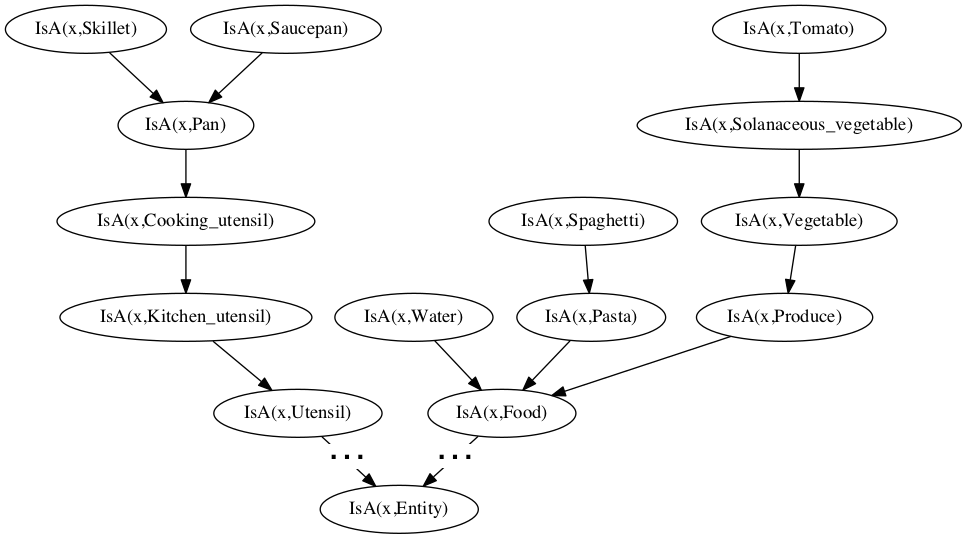}
\includegraphics[width=0.39\textwidth]{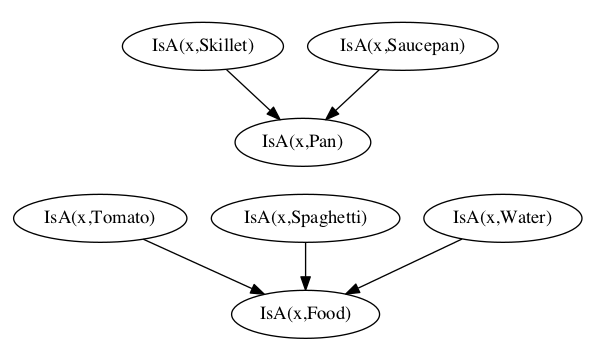}
\caption{An example of the \emph{IsA} relation before (left) and after (right) compression.  Compression reduces the amount of redundant and high-level information in the network. }
\label{Compression}
\end{figure*}

\subsection{Seed Word Disambiguation}
After the seed words have been collected, they must be disambiguated to determine the contextually correct senses of the words.  For example, the word "pan" has the following four senses in WordNet:

\begin{enumerate}
\item pan, cooking pan -- cooking utensil consisting of wide metal vessel
\item Pan, goat god -- (Greek mythology) god of fields and woods and shepherds and flocks
\item pan -- shallow container made of metal
\item Pan, genus Pan -- chimpanzees; more closely related to Australopithecus than to other pongids
\end{enumerate}
Given a particular environment, not all of the above senses will be contextually relevant.  To keep the size of the network small and contextually accurate, the seed words can be disambiguated and the irrelevant senses can be excluded from the generated network.

Since WordNet provides information on the different word senses, it can be used to perform this disambiguation.  The approach used for this paper is similar to that in \cite{Tsatsaronis2008WordNetworks}.  Given that the seed words originate from the same context, they are likely to be semantically similar.  Therefore, disambiguation can be performed by finding the sense of each word that maximizes the overall similarity between the seed words.  To do so, the disambiguation algorithm finds a Minimum Spanning Tree (MST), where each node represents the most relevant sense of one of the seed words.  
% Explain that you are using an MST and why you are using it
% - Since seed words are likely to be semantically similar, we attempt to find the
%	senses of each word that maximize the similarity between seed words
% - Find an MST where each node is the chosen sense of one of the seed words
Given a set of $n$ seed words, $W = \{w_1, w_2, ... , w_n\}$, and a set of possible senses (synset), $S_i = \{s_i^1, s_i^2, ... \}, i \in \{1, 2, ... , n\}$, for each seed word, the MST, $T = \{s_1^*, s_2^*, ... , s_n^*\}$, is computed, where $s_i^*$ indicates the chosen sense of the word, $w_i$, that minimizes the cost of the tree.  The cost metric used for this algorithm is $C_{ij} = \min_{k,l}1 - wup(s_i^k, s_j^l)$, where $wup()$ is the Wu \& Palmer similarity measure \cite{Wu1994VerbsSelection} for the senses, $k$ and $l$, of the $i^{th}$ and $j^{th}$ words, respectively.  This measure is based on the length of the path between the two senses of the two words in the WordNet hierarchy, where a longer path generally indicates less semantic similarity.

Since it is possible for a given set of seed words to have multiple minimum spanning trees, the starting node for the MST is chosen as the word, $w_i$, for which the number of senses, $|S_i|$, is a minimum.  Then for each of the senses of this word, the MST is computed, and the MST with the lowest overall cost, determined by the sum of the costs of all edges in the MST, is chosen.  While this does not guarantee that the best possible MST is found, it avoids having to compute all possible MSTs, thereby reducing computation time.  Using the MST approach also assumes that all of the seed words are connected.  While this might not always be true, it is likely that the seed words are related by context, so the MST approach should yield good results in most cases.

\subsection{IsA Relation and Compression}
After the seed words have been disambiguated, the \emph{IsA} relation is added to the network by traversing the WordNet hypernym hierarchy from each of the disambiguated seed words to the root node and adding each node along this path to the network.  Although WordNet is hand-coded, it does contain a large amount of redundant and high-level concepts that convey little information, as shown in Figure \ref{Compression}.  If these nodes are not removed from the network, this can lead to rapid expansion in the size of the network when other relations are added.  To reduce the size of the network to a manageable level and remove the high-level and redundant nodes, the compression strategy implemented in \cite{Stoica2004Nearly-AutomatedCreation} is employed.  The compression uses the following three rules:

\begin{enumerate}
\item Eliminate selected top-level (very general) categories, like abstraction, entity.
\item Starting from the leaves, eliminate a parent that has fewer than n children, unless the parent is the root.
\item Eliminate a child whose name appears within the parent's.
\end{enumerate}

For the first rule, "top-level" categories are defined as words with an Information Content (IC) of less than 5.0 when evaluated against the Brown corpus.
% IC(concept) = -log(P(concept)) P(concept) in corpus
% http://citeseerx.ist.psu.edu/viewdoc/download;jsessionid=B1F9AAB6BE8418A7800B57FE01CD4F1E?doi=10.1.1.59.2199&rep=rep1&type=pdf

\begin{figure*}
\centering
\includegraphics[width=\textwidth]{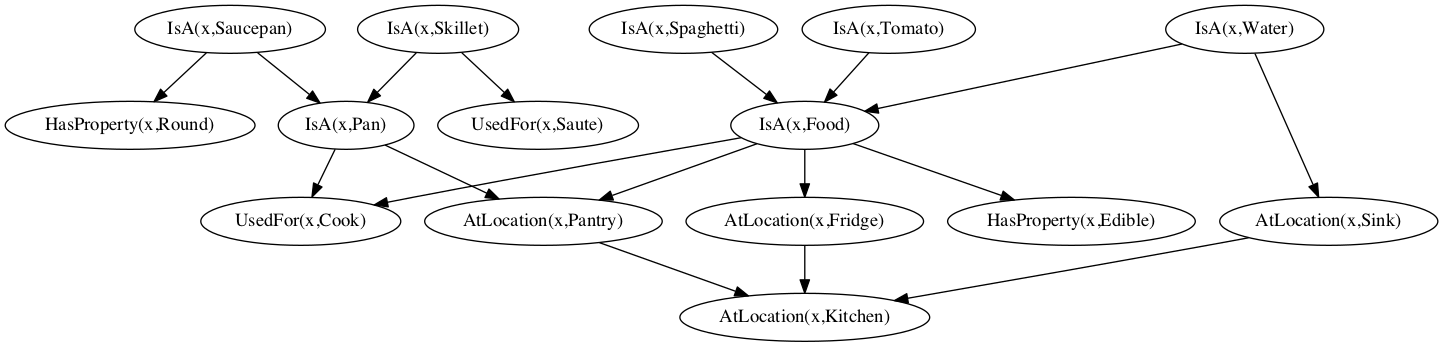}
\caption{An example of a network after adding all four types of relations from WordNet and ConceptNet.}
\label{NetFinal}
\end{figure*}

\subsection{Adding ConceptNet Relations}
In addition to the \emph{IsA} relation, the \emph{UsedFor}, \emph{HasProperty}, and \emph{AtLocation} relations are added from ConceptNet.  To do so, the relations in ConceptNet are first disambiguated to remove contextually irrelevant relations.  An approach similar to that in \cite{Chen2011CombiningDisambiguation} is used for this purpose.  For each ConceptNet relation, \emph{<c, relation, d>}, where $d$ is an ambiguous word and $c$ is disambiguated, the Word Sense Profile, $WSP(d_i) = {w_1, w_2, ... }$ is generated for each sense, $d_i$, of the word, $d$.   Each $w_j$ in the WSP is a word from one of the following sources in  WordNet:

\begin{enumerate}
\item All synonyms of $d_i$
\item All words (excluding stop words) in the gloss/definition for $d_i$
\item All direct hypernyms (parent nodes) and hyponyms (child nodes) of $d_i$ in WordNet
\item All meronyms/holonyms (has part or part of) relations in WordNet
\item All words (excluding stop words) in the glosses of the direct hyponyms of $d_i$
\end{enumerate}

After the WSP has been generated for each sense, a score is computed for each of the WSPs.  This score is equal to the sum of the semantic relatedness between the non-ambiguous word, $c$, and each word in $WSP(d_i)$.  The $i^{th}$ sense is chosen if the score is maximal for that value of $i$.  In \cite{Chen2011CombiningDisambiguation}, the relatedness is measured using the Normalized Google Distance (NGD), which is based on the number of Google hits returned for the two words together.  However, since the current version of the Google Search API limits the number of queries per day, a different semantic relatedness measure called Explicit Semantic Analysis (ESA) \cite{Gabrilovich2007ComputingAnalysis} was used instead.  ESA uses a pre-processed dump of Wikipedia to generate a large table, where the columns, $C = \{c_i\}, i \in [1, N]$, represent each of the concepts (pages) in Wikipedia, and the rows, $W = \{w_j\}, j \in [1, M]$, represent the words on those pages.  The entries, $E = \{e_{ij}\}$, in the table represent the frequency count of the words, $w_j$, in each Wikipedia page, $c_i$.  The semantic relatedness between two words is computed by taking each corresponding row, $E_j$, as a weighted vector of concepts and computing the cosine distance between the two vectors.

After each of the ConceptNet relations has been disambiguated, only the relations corresponding to the correct senses of each word are added to the BLN.  This helps prevent contextually irrelevant information from being added to the network.  Additionally,  relations, \emph{<c, relation, d>}, where $d$ consists of more than one word were excluded from the network.  Since ConceptNet is not hand-coded, it contains a significant amount of noisy or erroneous relations.  Excluding such relations was found to significantly reduce the size of the network without removing a large number of correct relations.

Due to the hierarchical nature of locations, two hops in ConceptNet were allowed when adding the \emph{AtLocation} relation to the network.  However, to prevent the size of the network from expanding rapidly with the increased number of hops, and to prevent contextually irrelevant locations from being added to the network, any locations that were added to the network were constrained to locations within the robot's current environment.  For example, \emph{IsA}(x,Food) $\rightarrow$ \emph{AtLocation}(x,Store), would be excluded if the robot's current environment is in the kitchen, since stores are not located within kitchens.  A sample of the output of the algorithm after adding the relations from ConceptNet can be seen in Figure \ref{NetFinal}.

\subsection{Weight Learning}
The final component of the network generation is to learn the CPF for each fragment in the network.  To do so, a set of training evidence is generated with a likelihood equal to a linear combination of the weights assigned to each relation in ConceptNet and the ESA relatedness measure between the two concepts in the relation.  In future work, this evidence will be augmented with evidence collected by the robot, but generating simulated evidence will provide an initial estimate of the real-world probabilities and enable inference results to be ranked according to their relative likelihoods.  Once the evidence has been collected, the CPFs can be learned via maximum likelihood by counting the frequency of each child node being true for each configuration of the parent nodes.

\section{Evaluation}
To evaluate the network generation algorithm, three sets of seed words were collected, from three different task scenarios related to typical household chores.  The three scenarios included cooking a recipe, doing laundry, and cleaning the house, with 19, 15, and 11 seed words, respectively.  An example of some of the relations the network generated in each case can be seen in Table \ref{Output}.  For each seed word, $s_i$, inference was run over the network where the evidence variable was IsA(Object\_i, $s_i$) with a value set to true.  Queries were then made for the variables IsA(Object\_i, x), AtLocation(Object\_i, x), HasProperty(Object\_i, x), AtLocation(Object\_i, x), and UsedFor(Object\_i, x) for each seed word.  The output of the inference process was then compared to a gold standard.  This gold standard was created by hand labeling each of the possible inference outputs as either true or false.  Each query result was assumed to be true if the associated probability was greater than 0.5, and false otherwise.  Table \ref{Results} shows the results from each of the three scenarios on each of the four query types.
\begin{table}
\centering
\bgroup
\def\arraystretch{1.25}
\begin{tabular}{| c || p{6.25cm} |}
\hline
Source & Relations\\
\hline
Recipe & IsA(x, Garlic) $\rightarrow$ IsA(x, Flavorer) \newline IsA(x, Food) $\rightarrow$ AtLocation(x, Container) \newline IsA(x, Container) $\rightarrow$ HasProperty(x, Plastic) \newline IsA(x, Stove) $\rightarrow$ UsedFor(x, Heat)\\
\hline
Laundry & IsA(x, Washer) $\rightarrow$ IsA(x, Appliance) \newline IsA(x, Sock) $\rightarrow$ AtLocation(x, Dresser) \newline IsA(x, Towel) $\rightarrow$ HasProperty(x, Cotton) \newline IsA(x, Shirt) $\rightarrow$ UsedFor(x, Dress)\\
\hline
Cleaning & IsA(x, Rag) $\rightarrow$ IsA(x, Piece\_of\_cloth) \newline IsA(x, Soap) $\rightarrow$ AtLocation(x, Sink) \newline IsA(x, Paper\_towel) $\rightarrow$ HasProperty(x, Paper) \newline IsA(x, Broom) $\rightarrow$ UsedFor(x, Sweep)\\
\hline
\end{tabular}
\egroup
\caption{An example of relations added to each of the three test networks.}
\label{Output}
\end{table}

In all three cases, the highest accuracy occurred for the \emph{IsA} relation.  Since WordNet is hand-coded, the majority of the \emph{IsA} relations were correct when compared to the gold standard.  Most of the errors that occurred with the \emph{IsA} relation corresponded to the seed words that had not been disambiguated correctly.  As shown in Table \ref{WSD}, the lowest disambiguation accuracy occurs for the recipe scenario, and the highest accuracy is achieved for the cleaning scenario.  This corresponds to the inference performance for each of the three tasks, with recipe achieving the lowest accuracy and cleaning achieving the highest.

Overall, accuracies for the \emph{AtLocation}, \emph{HasProperty}, and \emph{UsedFor} relations ranged from the low to mid seventies to upper eighties.  Several sources of error limited the inference accuracy of the network.  Errors made during earlier stages in the network tended to propagate through the rest of the network.  For example, if a seed word was incorrectly disambiguated, the relations added from ConceptNet would often be related to the incorrect sense of the seed word.  Other sources of error came from the noise present in ConceptNet.  In some cases, the relations themselves are inaccurate -- the recipe network, for example, included the relation IsA(x, Container) $\rightarrow$ UsedFor(x, Wash).  In other cases, this noise appeared in the form of missing connections within the network.  Although the IsA(x, Saucepan) $\rightarrow$ UsedFor(x, Saute) connection existed in the recipe network, the IsA(x, Frying\_pan) $\rightarrow$ UsedFor(x, Saute) relation was missing, though both objects are arguably equally suited to the task of sauteing.  This error occurred because the IsA(x, Frying\_pan) $\rightarrow$ UsedFor(x, Saute) does not exist at all in ConceptNet.  

\section{Future Work}
One of the main goals of future work is to perform more extensive evaluation on the network generation algorithm.  This could include use of crowdsourcing to develop a gold standard that more accurately reflects the uncertainty of the relations.  Additionally, the algorithm could be tested across several different domains to determine whether it generalizes beyond the household scenarios presented in this paper.

Another future goal is to perform grounding to allow a robot to associate the real objects it encounters in its environment with the abstract concepts over which it can perform inference.  Grounding the network will enable the robot to use this high-level knowledge to perform plan repair by locating missing objects or finding suitable substitutes.  At the point where task execution fails, the appropriate query can be formulated, and the ranked inference results can then be grounded to allow the robot to attempt to continue execution.

The last goal of this work is to enable both the network structure and associated probability distribution to be updated online without the need to regenerate the network.  Updates to the conditional probabilities could be performed as the robot collects new evidence on its own or through interaction with humans such as question asking.  The structure of the network could also be modified by adding or removing nodes as the robot encounters new objects or determines that portions of the network have been unused and are unnecessary.

\begin{table}
\centering
\bgroup
\def\arraystretch{1.25}
\begin{tabular}{| c || c | c | c | c |}
\hline
Source & IsA & AtLocation & HasProperty & UsedFor\\
\hline
Recipe & 97.6 & 86.8 & 82.0 & 88.1\\
Laundry & 98.3 & 77.3 & 88.9 & 89.5\\
Cleaning & 98.6 & 72.7 & 94.7 & 79.2\\
\hline
\end{tabular}
\egroup
\caption{Total inference accuracy (as a percentage) over the set of seed words from each source when compared to the gold standard.}
\label{Results}
\end{table}

\begin{table}
\centering
\bgroup
\def\arraystretch{1.25}
\begin{tabular}{| c || c |}
\hline
Source & Accuracy\\
\hline
Recipe & 73.7\\
Laundry & 80.0\\
Cleaning & 81.8\\
\hline
\textbf{Mean} & 78.5\\
\hline
\end{tabular}
\egroup
\caption{Seed word disambiguation accuracy for each of the three testing scenarios.}
\label{WSD}
\end{table}

% Refs
\bibliography{Mendeley}
\bibliographystyle{aaai}
\end{document}